\documentclass{article}
\usepackage{amsmath,graphicx,mlspconf}
\usepackage{xcolor}
\usepackage{color, colortbl}
\usepackage{amsmath}
\usepackage{amssymb}
\usepackage{tikz}
\usetikzlibrary{positioning, arrows.meta, shapes.geometric}
\usepackage{hyperref}
\hypersetup{
    colorlinks,
    linkcolor={blue!80!black},
    citecolor={blue!80!black}
}

\title{Isolating Nonlinear Independent Sources in fMRI with $\beta$-TCVAE Models}

\name{%
    Qiang Li$^{1}$\quad Shujian Yu$^{2}$\quad Jesus Malo$^{3}$\quad Jingyu Liu$^{1,4}$\quad T{\"u}lay~Adali$^{5}$\quad Vince D. Calhoun$^{1,4}$\thanks{This work was supported by the National Science Foundation (NSF) grant 2112455, 2316420, 2316421 and the National Institutes of Health (NIH) grants R01MH123610 and R01MH119251.}%
}
\address{%
    $^{1}$Tri-Institutional Center for Translational Research in Neuroimaging and Data Science (TReNDS),\\
    Georgia State University, Georgia Institute of Technology, and Emory University, Atlanta, GA, USA \\
    $^{2}$Department of Computer Science, Vrije Universiteit Amsterdam, Amsterdam, the Netherlands \\
    $^{3}$Image Processing Laboratory, University of Valencia, Valencia, Spain \\
    $^{4}$Department of Computer Science, Georgia State University, Atlanta, GA, USA\\ 
    $^{5}$Department of Computer Science and Electrical Engineering,\\ University of Maryland, Baltimore County, Baltimore, MD, USA%
}
\begin{document}
\maketitle
\begin{abstract}
Learning meaningful latent representations from nonlinear fMRI data remains a fundamental challenge in neuroimaging analysis. Traditional independent component analysis, widely used due to its ability to estimate interpretable functional brain networks, relies on a linear mixing assumption for latent sources, limiting its ability to capture the inherently nonlinear and complex organization of brain dynamics. More recently, deep representation learning methods have emerged as promising alternatives for modeling nonlinear latent structure. However, many of these approaches have been evaluated primarily on simulated datasets or natural image benchmarks, with comparatively limited validation on real-world neuroimaging data such as fMRI. In this work, we are motivated by the $\beta$-TCVAE (Total Correlation Variational Autoencoder), a refinement of the $\beta$-VAE framework for learning latent representations without introducing additional hyperparameters during training. We adapt and modify this model to fMRI data for nonlinear source disentanglement, aiming to separate mixed spatial and temporal brain signals into interpretable components. We show that the $\beta$-TCVAE framework can recover meaningful nonlinear spatial components with biological relevance, including well-established intrinsic connectivity networks such as the default mode network. Furthermore, we evaluate the learned representations using functional network connectivity, showing that the latent structure captures coherent and interpretable brain organization patterns. This study provides a pilot investigation that bridges nonlinear representation learning and fMRI analysis. 
\end{abstract}


\section{Introduction}
\label{sec:intro}
The blind source separation (BSS) problem is a fundamental problem in signal processing, where the goal is to recover underlying source signals from observed mixtures without prior knowledge of the mixing process~\cite{choi2005blind,comon2010handbook,lee1997blind,amari1997blind}. Independent component analysis (ICA)~\cite{hyvarinen2000independent,hyvarinen1999fast,bell1995information} is a widely used computational method to address this problem, with applications across various fields for extracting statistically independent components from mixed signals.

In the neuroimaging domain, ICA is widely used to identify intrinsic connectivity networks (ICNs) from functional MRI (fMRI) data~\cite{calhoun2009review,calhoun2002method}. Group ICA~\cite{calhoun2002method} enables estimation of common network structure across subjects while also providing subject-specific spatial maps and time courses through back-reconstruction procedures. Each ICN can then be mapped to functional brain network domains, providing an interpretable representation of large-scale brain organization. More recently, spatially constrained ICA approaches~\cite{du2020neuromark,hesse2005fastica} have extended this framework by incorporating reliable spatial priors to improve component correspondence, reproducibility, and robustness across datasets and individuals. All these methods facilitate downstream analyses such as estimation of functional network connectivity (FNC), helping to better characterize interactions among brain networks and identify abnormal connectivity patterns associated with neurological and psychiatric disorders~\cite{li2026spatiotemporal,li2025higher}.

However, traditional ICA does not model nonlinear, hierarchical, or temporally adaptive structure. In fact, modern extensions such as constrained ICA~\cite{du2020neuromark,hesse2005fastica}, tensor ICA~\cite{virta2016applying}, and nonlinear ICA~\cite{khemakhem2020variational,Malo06ica,hyvarinen2019nonlinear,hyvarinen2016unsupervised,almeida2003misep,li2025deep} have substantially relaxed many of these limitations. However, most of these advances have been evaluated primarily on simulated data or natural image benchmarks to demonstrate the performance of the proposed frameworks. In contrast, there is still limited application and systematic validation in neuroimaging analysis. This gap highlights the need to further explore and adapt these methods for real brain imaging data, where complex spatial-temporal dependencies and biological variability present additional challenges.

In this work, we are motivated by the $\beta$-TCVAE (total correlation variational autoencoder)~\cite{chen2018isolating}, which is an extension of the $\beta$-VAE~\cite{kingma2013auto} that explicitly encourages disentangled representations by penalizing the total correlation (TC) among latent variables. We adapt and modify this framework to better handle large-scale fMRI datasets, enabling it not only to extract meaningful spatial ICNs, but also to construct corresponding time series that are aligned with each spatial component. We evaluate the proposed method on real fMRI data, and the results demonstrate that the adapted $\beta$-TCVAE performs well in this initial pilot study.

\section{Methods}
\label{sec:met}
\subsection{Model Overview}
We consider multi-subject fMRI data $\{x_t^{(s)} \in \mathbb{R}^D\}$, where $s \in \{1,\dots,S\}$ indexes subjects and $t$ indexes time. Our goal is to learn a low-dimensional latent representation $z \in \mathbb{R}^K$ that captures shared latent structure across subjects while remaining disentangled. We model the data using a subject-conditioned variational autoencoder of the form
\begin{equation}
q_\phi(z \mid x, s), \quad p_\theta(x \mid z),
\end{equation}
where $q_\phi(z \mid x, s)$ is the encoder that infers a latent distribution $z$ from input $x$ and subject identity $s$, and $p_\theta(x \mid z)$ is the decoder that reconstructs $x$ from $z$. The parameters $\phi$ and $\theta$ denote the encoder and decoder networks, respectively.

\subsection{Subject-Conditioned Variational Model}
To incorporate inter-subject variability, each subject index $s$ is mapped to a learned embedding
\begin{equation}
e_s = E_\psi(s) \in \mathbb{R}^m,
\end{equation}
where $E_\psi(\cdot)$ is an embedding lookup table parameterized by $\psi$, and $e_s$ represents a learned vector capturing subject-specific effects.

The subject embedding is concatenated with the input as $\tilde{x} = [x; e_s]$, forming a subject-conditioned representation that is fed into the encoder. The variational posterior is defined as a diagonal Gaussian:
\begin{equation}
q_\phi(z \mid x, s) = \mathcal{N}(\mu_\phi(\tilde{x}), \mathrm{diag}(\sigma_\phi^2(\tilde{x}))).
\end{equation}
Here, $\mu_\phi(\tilde{x}) \in \mathbb{R}^K$ and $\sigma_\phi^2(\tilde{x}) \in \mathbb{R}^K$ are the outputs of the encoder network parameterized by $\phi$, representing the mean and variance of the latent distribution, respectively.

Latent variables are sampled using the reparameterization trick,
\begin{equation}
z = \mu + \sigma \odot \epsilon, \quad \epsilon \sim \mathcal{N}(0, I),
\end{equation}
where $\epsilon$ is an auxiliary noise variable enabling backpropagation through stochastic sampling, and $\odot$ denotes element-wise multiplication.

The generative model defines the likelihood of the observed data as
\begin{equation}
p_\theta(x \mid z) = \mathcal{N}(\hat{x}_\theta(z), I),
\end{equation}
where $\hat{x}_\theta(z)$ is the decoder output parameterized by $\theta$, representing the reconstructed fMRI signal from the latent representation.

\subsection{$\beta$-TCVAE Objective}
We optimize a decomposition of the ELBO into reconstruction, mutual information (MI), TC, and dimension-wise Kullback-Leibler (KL) terms:
\begin{equation}
\mathcal{L} =
\mathcal{L}_{\mathrm{rec}} +
\mathrm{MI} +
\beta \, \mathrm{TC} +
\mathrm{KL}_{\mathrm{dim}}.
\end{equation}

The reconstruction loss corresponds to a Gaussian likelihood:
\begin{equation}
p_\theta(x \mid z) = \mathcal{N}(\hat{x}_\theta(z), \sigma^2 I),
\end{equation}
leading to:
\begin{equation}
\mathcal{L}_{\mathrm{rec}} = \mathbb{E}_{q_\phi(z|x)} \|x - \hat{x}_\theta(z)\|_2^2.
\end{equation}

The aggregated posterior is defined as:
\begin{equation}
q(z) = \int q(z \mid x) p(x)\, dx,
\end{equation}
where $q(z)$ represents the marginal distribution of latent variables induced by the data distribution $p(x)$ and the variational posterior $q(z|x)$.

MI between observed data $x$ and latent variables $z$ is given by:
\begin{equation}
\mathrm{MI} =
\mathbb{E}_{q(x,z)} \left[
\log q(z \mid x) - \log q(z)
\right],
\end{equation}
where $q(x,z)=p(x)q(z|x)$ denotes the joint distribution induced by the model.

TC measures the dependence between latent dimensions and is defined as:
\begin{equation}
\mathrm{TC} =
\mathbb{E}_{q(z)} \left[
\log q(z) - \sum_{j=1}^K \log q(z_j)
\right],
\end{equation}
where $q(z_j)$ denotes the marginal distribution of the $j$-th latent dimension.

Dimension-wise KL divergence is given by:
\begin{equation}
\mathrm{KL}_{\mathrm{dim}} =
\sum_{j=1}^K
\mathbb{E}_{q(z_j)}
\left[
\log q(z_j) - \log p(z_j)
\right],
\quad p(z)=\mathcal{N}(0,I),
\end{equation}
where $p(z)$ is an isotropic standard normal prior over latent variables.

Overall, this objective promotes latent representations that capture structured fMRI patterns while encouraging statistical independence across latent dimensions, enabling more interpretable components for factorizing brain activity.

\subsection{$\beta$-TCVAE Architecture and Training Details}
The model is a subject-conditioned variational autoencoder consisting of a multilayer perceptron encoder and decoder. The encoder receives as input a concatenation of the fMRI feature vector and a learned subject embedding of dimension 8. This subject embedding is implemented as a trainable lookup table and is jointly optimized with the rest of the network, allowing the model to capture subject-specific variability.

The encoder is composed of two fully connected hidden layers with 512 and 256 units, respectively, each followed by ReLU activations. The decoder mirrors this structure with two hidden layers of 256 and 512 units and ReLU activations, mapping latent variables back to the original fMRI feature space to reconstruct the input signal.

The model is trained using the Adam optimizer~\cite{kingma2014adam} with a learning rate of $1\times10^{-4}$ and a batch size of 64 for 16000 epochs. Gradient norms are clipped to a maximum value of 5.0 to stabilize training. A linear warm-up schedule is applied to the $\beta$ coefficient in the $\beta$-TCVAE objective over the first 10 epochs, after which it is kept fixed. The latent dimensionality is set to 80. Input fMRI data are standardized voxel time series extracted within a brain mask, optionally reduced using principal component analysis (PCA) to 100 dimensions prior to training. 

\subsection{Dataset}
The Human Connectome Project (HCP) dataset~\cite{van2013wu} used in this study underwent a standard preprocessing pipeline. Initial volumes were first discarded to ensure signal stabilization, followed by slice timing correction and rigid-body motion correction. These steps were performed using the FMRIB Software Library (FSL v6.0)~\cite{jenkinson2012fsl} and the Statistical Parametric Mapping toolbox (SPM12)~\cite{friston2007short} in MATLAB (R2020a). 

Subsequently, the functional images were registered to the Montreal Neurological Institute (MNI) template to enable normalization to a standard anatomical space. Finally, the data were resampled to 3 mm isotropic voxels and spatially smoothed using a Gaussian kernel with 6 mm full width at half maximum (FWHM).

\section{Results}
\label{sec:res}
\subsection{Disentangling Spatial and Temporal Components via $\beta$-TCVAE in fMRI Data}
\begin{figure*}[!ht]
    \centering
    \includegraphics[width=\linewidth, height=12cm]{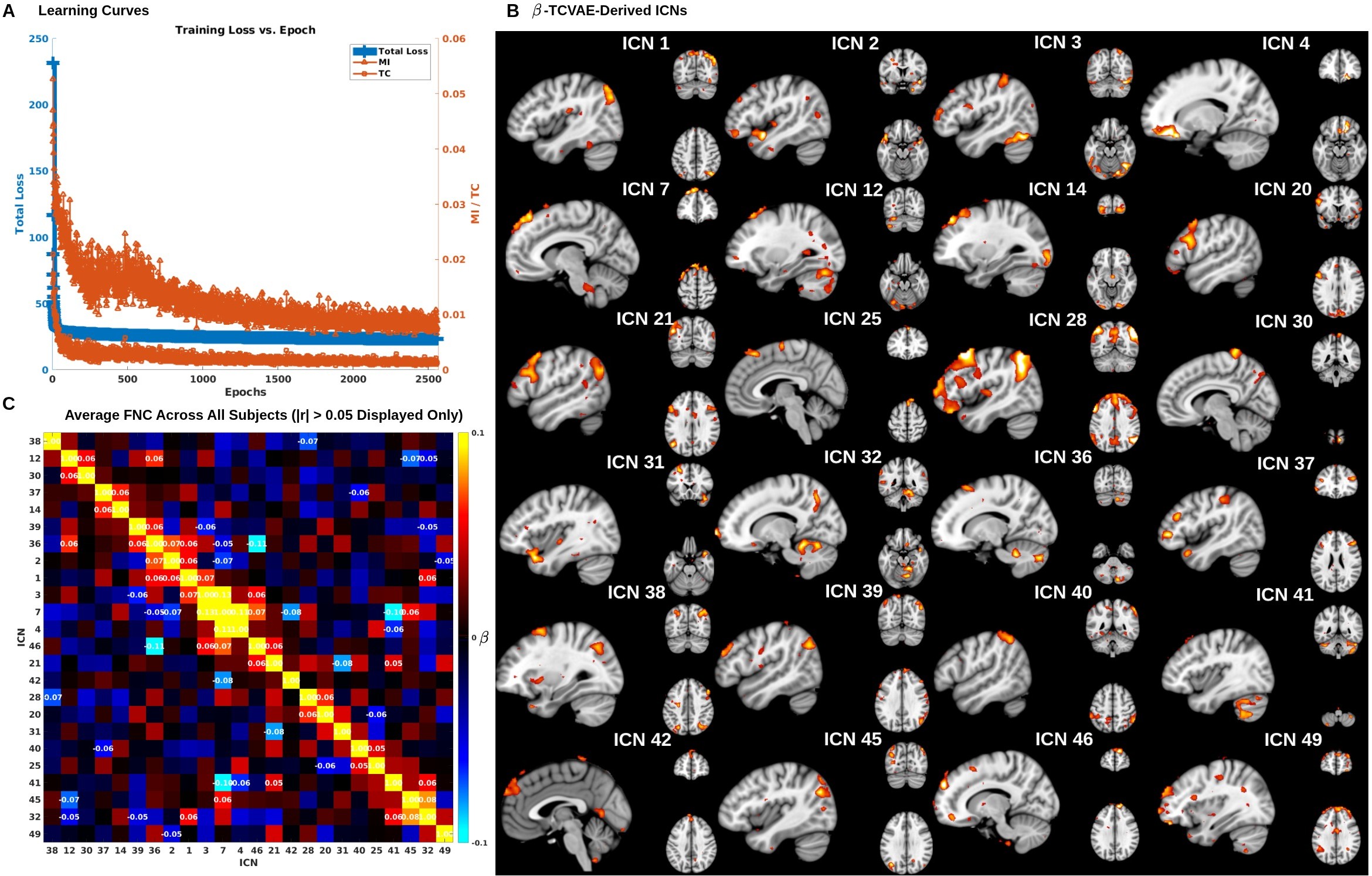}
    \caption{Overview of the $\beta$-TCVAE framework for disentangled nonlinear ICA in fMRI analysis. \textbf{A.} Training curves showing the total loss, MI, and TC during optimization. \textbf{B.} ICNs identified by the $\beta$-TCVAE framework. \textbf{C.} Average FNC across all HCP subjects, showing only correlations with $|r| > 0.05$.}
    \label{fig:1}
\end{figure*}

We successfully disentangled spatial components and temporal dynamics in fMRI data using a $\beta$-TCVAE framework. Fig.\ref{fig:1}A illustrates the training dynamics of the $\beta$-TCVAE framework, including the total loss, MI, and TC. As training progresses, the total loss steadily decreases, while both MI and TC are progressively minimized. This behavior suggests improved disentanglement and separation of latent components over time. Additional training epochs are expected to further refine the nonlinear source separation process.

Fig.\ref{fig:1}B presents representative ICNs identified by the $\beta$-TCVAE model. Several components exhibit clear multi-network coupling patterns, indicating that the learned latent representations capture higher-order dependencies across brain networks. For example, ICN 28 strongly corresponds to the default mode network (DMN), one of the core brain networks observed in resting-state fMRI. The results suggest that a disentanglement-driven nonlinear ICA framework can naturally recover meaningful multi-network coupling interactions in fMRI data.

Fig.\ref{fig:1}C shows the average FNC matrix computed across all HCP subjects, displaying only correlations with $|r| > 0.05$. The FNC matrix was reordered using hierarchical clustering to group similar components together. Although the analysis includes 24 components, distinct clustered structures are already visible along the diagonal, suggesting organized interactions among subsets of networks. These clustering patterns are expected to become more pronounced with a larger number of components.

\subsection{Spatial Components from $\beta$-TCVAE Compared with InfoMax}
\begin{figure*}[!ht]
    \centering
    \includegraphics[width=\linewidth, height=13.8cm]{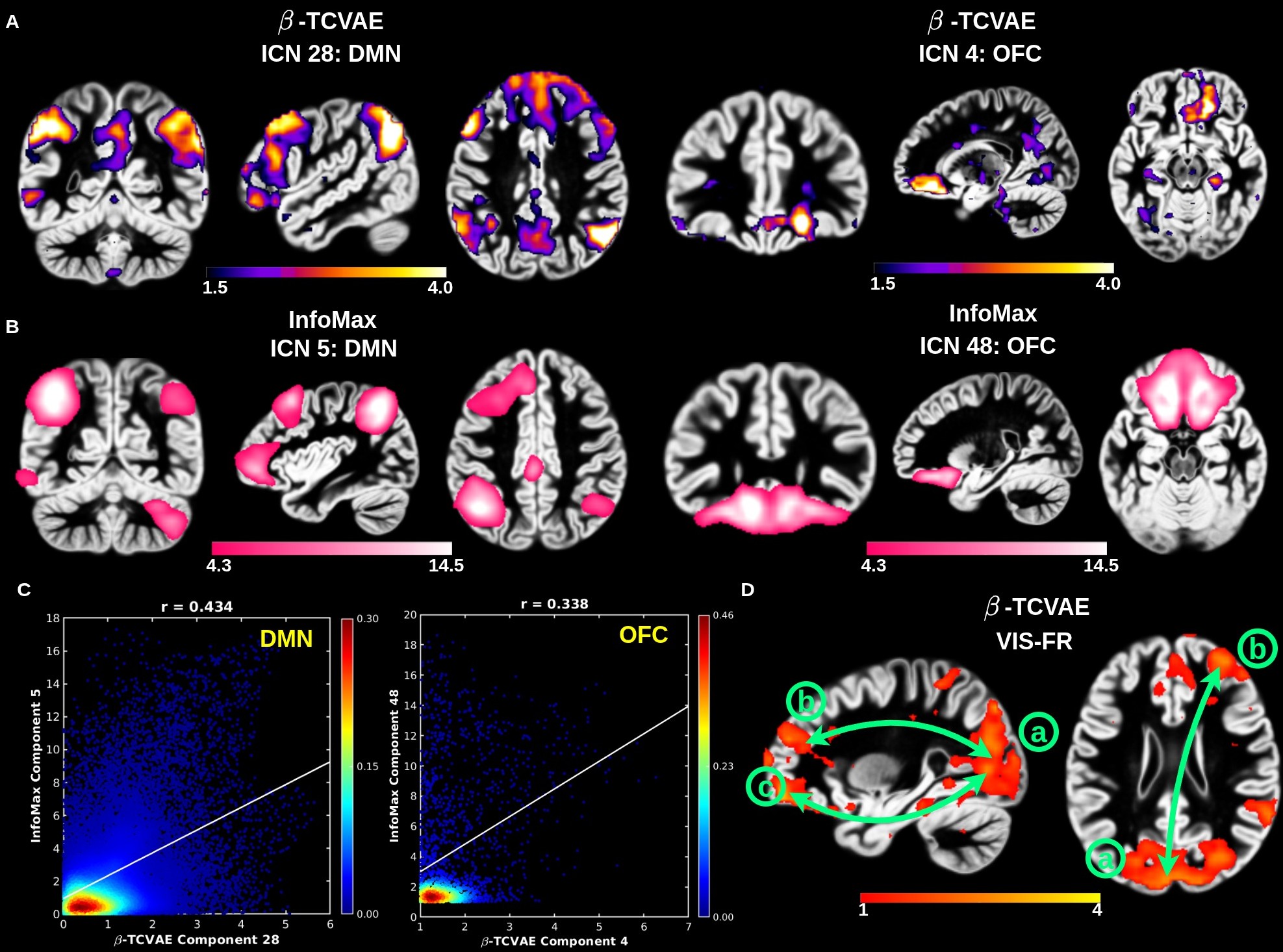}
    \caption{Representative ICNs identified by $\beta$-TCVAE and InfoMax. \textbf{A.} DMN and OFC components derived from $\beta$-TCVAE (colorbar indicates voxel intensity). \textbf{B.} DMN and OFC components derived from InfoMax. \textbf{C.} Spatial component correlations of the DMN and OFC from $\beta$-TCVAE and InfoMax, respectively. \textbf{D.} Unique spatial components identified by $\beta$-TCVAE, highlighting its ability to simultaneously capture two ICNs (\textcircled{a}-VIS, \textcircled{b}/\textcircled{c}-FR).}
    \label{fig:2}
\end{figure*}

Representative ICNs derived from $\beta$-TCVAE and InfoMax are shown in Fig.\ref{fig:2}. The $\beta$-TCVAE framework identified well-defined resting-state networks, including the DMN (ICN 28) and orbitofrontal cortex (OFC; ICN 4), as illustrated in Fig.\ref{fig:2}A. The extracted DMN component exhibited clear spatial coverage across canonical DMN regions, including the posterior cingulate cortex (PCC), medial prefrontal cortex (mPFC), and lateral parietal regions, demonstrating that the model effectively captured the core functional organization of the resting-state brain.

In comparison, matched components obtained from the InfoMax ICA approach~\cite{bell1995information}, specifically ICN 5 (DMN) and ICN 48 (OFC), showed relatively less complete spatial representations, as shown in Fig.\ref{fig:2}B. In particular, the DMN extracted by InfoMax exhibited reduced spatial continuity across several canonical DMN regions. In contrast, the $\beta$-TCVAE-derived DMN appeared more spatially coherent and more consistent with established resting-state network patterns.

To quantitatively evaluate the similarity between the two methods, spatial correlation analyses were performed between corresponding ICNs derived from $\beta$-TCVAE and InfoMax, as presented in Fig.\ref{fig:2}C. The resulting correlations were moderate, indicating that both approaches capture comparable functional network structures. However, despite these overall similarities, the $\beta$-TCVAE components consistently exhibited clearer boundaries and more complete spatial coverage. These findings suggest that $\beta$-TCVAE provides more robust and interpretable spatial representations of resting-state ICNs compared with conventional InfoMax ICA.

To further highlight the advantages of $\beta$-TCVAE, representative components are presented in Fig.\ref{fig:2}D, showing that the model can simultaneously capture multiple functional networks, including the visual (VIS) and frontal (FR) networks. Such coupled or co-expressed components were not observed in the InfoMax results. This suggests that $\beta$-TCVAE may better preserve interactions between distributed brain networks, whereas InfoMax tends to separate these patterns into independent components, potentially overlooking multi-network coupling effects.

\section{Discussion}
\label{sec:dis}
Our results demonstrate that $\beta$-TCVAE can effectively learn biologically meaningful spatial components that are well aligned with canonical brain networks. Compared with linear ICA (InfoMax)~\cite{bell1995information}, the learned representations are more spatially coherent and exhibit improved correspondence with established functional networks.

Importantly, $\beta$-TCVAE also captures patterns of multi-network co-activation that are not explicitly represented in InfoMax-derived components. These coupled spatial patterns suggest that the model may better preserve interactions between distributed brain networks, whereas traditional ICA tends to enforce stronger independence assumptions, which can separate or overlook such cross-network dependencies. This difference may be particularly relevant for understanding coordinated brain activity and functional integration.

One notable direction emerging from this study is the potential of $\beta$-TCVAE for scalable modeling of high-dimensional fMRI data. Future work could explore training on larger and more diverse cohorts to further improve the stability and generalizability of learned spatial representations. In addition, extending the framework to enable inference on unseen subjects could facilitate the extraction of consistent reference components across datasets. Finally, incorporating explicit spatial constraints may further enhance the biological plausibility and interpretability of the learned networks.

\section{Conclusion}
\label{sec:con}
In this pilot investigation, we adapted and modified the $\beta$-TCVAE framework to disentangle nonlinear representations in fMRI data. The results show that $\beta$-TCVAE can effectively extract biologically meaningful spatial components. In addition, we obtained corresponding time series associated with each spatial component, enabling further subject-level and network-level analysis. Overall, these results suggest that spatially structured, biologically meaningful components can be achieved within this framework, providing a basis for further development of constrained and interpretable representation learning methods for fMRI data, and extending traditional ICA toward representation learning approaches that better capture hierarchical and nonlinear brain dynamics.

\bibliographystyle{IEEEbib}
\bibliography{refs}

\end{document}